\documentclass[preprint,authoryear,12pt]{elsarticle}

\usepackage{lineno}

\usepackage{graphicx} 
\usepackage{subfigure}
\usepackage{epstopdf}
\usepackage{stfloats}
\usepackage{bm}
\usepackage{setspace}
\usepackage{tikz}
\usepackage{algorithm}
\usepackage{algpseudocode}
\usepackage{threeparttable}
\usepackage{amsmath,amssymb,amsfonts}

\usepackage{graphicx}
\usepackage{textcomp}

\usetikzlibrary{mindmap,backgrounds}
\usetikzlibrary{shapes,arrows,patterns} 

\begin{document}
\begin{frontmatter}

\title{Unknown Health States Recognition With Collective Decision Based Deep Learning Networks In Predictive Maintenance Applications}

\author[label1]{Chuyue Lou \corref{cor1}}
\ead{lcy2017@whut.edu.cn}
\author[label2]{M. Amine Atoui}
\ead{amine.atoui@gmail.com}

\cortext[cor1]{Corresponding author}

\address[label1]{School of Automation, Wuhan University of technology, China}
\address[label2]{Center for Applied Intelligent Systems Research, Halmstad University, Sweden}

\begin{abstract}
At present, decision making solutions developed based on deep learning (DL) models have received extensive attention in predictive maintenance (PM) applications along with the rapid improvement of computing power. Relying on the superior properties of shared weights and spatial pooling, Convolutional Neural Network (CNN) can learn effective representations of health states from industrial data. Many developed CNN-based schemes, such as advanced CNNs that introduce residual learning and multi-scale learning, have shown good performance in health state recognition tasks under the assumption that all the classes are known. However, these schemes have no ability to deal with new abnormal samples that belong to state classes not part of the training set. In this paper, a collective decision framework for different CNNs is proposed. It is based on a One-vs-Rest network (OVRN) to simultaneously achieve classification of known and unknown health states. OVRN learn state-specific discriminative features and enhance the ability to reject new abnormal samples incorporated to different CNNs. According to the validation results on the public dataset of Tennessee Eastman Process (TEP), the proposed CNN-based decision schemes incorporating OVRN have outstanding recognition ability for samples of unknown heath states, while maintaining satisfactory accuracy on known states. The results show that the new DL framework outperforms conventional CNNs, and the one based on residual and multi-scale learning has the best overall performance.
\end{abstract}

\begin{keyword}
convolutional neural network, one-vs-rest network, unknown heath states, classification, predictive maintenance
\end{keyword}
\end{frontmatter}

\section{Introduction}
\label{sec:introduction}

Modern industry has achieved a high degree of automation, and it is very important to ensure the long-term operation of industrial processes and the stability of product quality. However, many plant accidents have been reported despite being equipped with advanced process control systems, resulting in serious consequences including casualties, economic losses and environmental damage \citep{yuan2019hierarchical}. This is because the early detection of abnormal operating conditions and the execution of reliable decisions mainly rely on experienced operators, which are not always effective in preventing the occurrence and further deterioration of accidents \citep{wu2020self}. 

In industrial systems, predictive maintenance is critical in ensuring the safe and reliable operation of production processes \citep{LV2021101318}. Effective predictive maintenance systems provide early warning of the operating status of the system and equipment, which can avoid unexpected shutdown or even casualties due to various health states. In recent years, methods based on machine learning and statistical analysis have been developed and improved, enabling valuable information extraction and health state monitoring performance enhancement from massive process data. Principal Component Analysis (PCA), Partial Least Squares (PLS), Canonical Correlation Analysis (CCA), and Fisher Discriminant Analysis (FDA) have become widely employed feature extractors to obtain linear feature representations \citep{jiang2015performance, xu2017distributed, yin2014improved, xie2015advanced, huang2016canonical, jiang2015combined, wu2020data, wang2020novel}. As nonlinear feature extractors, kernel PCA, kernel FDA, kernel PLS, Support Vector Machine (SVM), and Artificial Neural Network (ANN) overcome the limitation of application in nonlinear systems \cite{hu2018fault, zhu2010fault, wang2016kernel, ren2016fault, malik2017emd}. In addition, dynamic PCA, dynamic PLS, Independent Component Analysis (ICA), and dynamic ICA have been used to describe other complex data characteristics, such as dynamic characteristics and non-Gaussian distribution assumptions \citep{rato2013fault, huang2015dynamic, lee2003multiple, li2017correlated}. The above traditional data-driven methods can be considered as shallow learning models, which have achieved good performance in many health state monitoring tasks. However, these shallow learning models may not be able to fully learn the complex mapping relationship between massive data and health states, resulting in insufficient generalization performance to big data \citep{lei2016intelligent}.

As one of the important tasks of predictive maintenance, heath states recognition aims to achieve fast and efficient discrimination between the system's faulty states. In the field of industrial system states recognition, Deep Learning (DL) models have received attention as an important branch of machine learning. Compared with shallow learning models, DL models perform well in learning more abstract and generalized latent representations under the influence of multi-layer nonlinear mappings. Also, DL model-based recognition methods tend to end-to-end learning, that is, integrating feature learning and recognition tasks as one training subject, which is more helpful for extracting state-related discriminative features. Convolutional Neural Network (CNN), Stacked Auto-encoder (SAE), Deep Belief Network (DBN), and Recurrent Neural Network (RNN) are the main families that have been widely employed and discussed in published research \citep{20202108704403, 20183205657811, 20194807748636, 20201008262884, 20201508392029, 20212010351759, 20212010350410, 20204909594579, cohen2022wavelet}. Relying on the superior properties of shared weights and spatial pooling, CNNs can learn effective representations of health states from industrial data, and have been developed and improved in many publications. In industrial processes, CNNs have been validated to have excellent recognition performance compared to other DL models \citep{wu2018deep, ge2021fault}. Considering the importance of limited labeled samples and sufficient unlabeled samples, Li et al. proposed a semi-supervised convolutional ladder network, which achieves the goal of extracting latent manifold features and enhancing incipient fault representations \citep{li2020semi}. In some scenarios, the mismatch of training and test sets also brings difficulties to the application of CNNs, which has been overcome and discussed in some studies. Li et al. integrated CNN and domain adaptation techniques including maximum mean discrepancy (MMD) and domain adversarial learning, reducing the mismatch between computer simulation and physical processes \citep{li2020transfer}. Wu and Zhao proposed a recognition model for multimode processes, which is based on joint MMD by aligning the joint distribution of multiple domain-specific layers \citep{wu2020fault}. Aiming to update the recognition model to include both existing heath states and new abnormal samples, Yu and Zhao proposed a broad CNN with incremental learning capability to avoid retraining processes \citep{yu2019broad}.

Most of the health states recognition methods based on DL models are based on the closed set assumption, in which all states are known a priori. Under this assumption, the features extracted by DL models from the training and test sets share the same class characterization \citep{he2021deep}. The health states recognition frameworks based on deep learning can achieve high accuracy with the support of complete datasets with guaranteed quantity and heath state diversity, but the acquisition cost is high. In practice, infrequent health states and expensive labeling costs make it difficult to identify and label all the states. In this scenario, new samples from unknown states will be misclassified into existing states. Therefore, it is necessary to consider samples of unknown states, which can be regarded as an important element for intelligent self-learning systems \citep{boult2019learning}. In industrial process health states recognition, several publications have suggested schemes to meet the needs of classifying known states while detecting unknown samples. Atoui et al. developed a single Conditional Gaussian Network with a modified distance criterion that allows simultaneous processing of known and unknown health states \citep{atoui2019single}. Considering the dynamic characteristics of industrial processes, Lou et al. further proposed an enhanced fault recognition method, which learns by extracting time-independent components instead of original data \citep{lou2020enhanced}. To overcome the limitation that common discriminative methods cannot identify unknown states, a modified discriminant rule is proposed to provide new statistical spaces for quadratic discriminant analysis (QDA), Fisher discriminant analysis (FDA), exponential discriminant analysis (EDA), orthonormal discriminant vector (ODV) and kernel Fisher discriminant (KFD)to ensure that each state is statistically isolated from other states \citep{lou2022novel}.

Most of the research on unknown health states recognition has focused on shallow learning methods. At present, almost all DL-based methods take the recognition of known states as the classification target. These DL methods set generalized boundaries to divide the closed decision space into subspaces for known states. They are effective under the assumption that all the new samples in the testing phase are part of the training set. But, they may fail in practice. Indeed, these closed set approaches may assign new samples from an unknown state to one of the states part of the training set. To overcome the above limitations, this paper proposes a new collective decision based framework for DL networks, which introduces one-vs-rest network (OVRN) as a classifier for CNN-based feature extractors. CNNs are responsible for extracting deep representations from input measurements, while OVRN further learns class-specific discriminative features and enhances the ability to reject samples of unknown health states. Furthermore, OVRN with a collective decision rule provides final recognition results through decision boundaries built for known health states. Experiments were conducted on the public dataset of Tennessee Eastman process (TEP), and the results show that the proposed CNN-based collective decision schemes have satisfactory accuracy in both known and unknown health states recognition. The main contributions of this work can be summarized as:
\begin{enumerate} 
	\item A new framework combining CNN-based feature extractors and an OVRN classifier is constructed, allowing deep learning methods like CNN and other models to process samples from known and unknown health states simultaneously.
	\item The features extracted by OVRN and the boundaries established by the collective decision rule are beneficial to the rejection of unknown samples and the preservation of recognition performance of known states.
	\item The public dataset from Tennessee Eastman Process (TEP) is used to verify the accuracy and effectiveness of proposed methods.
\end{enumerate}

The remainder of this paper is organized as follows. In Section 2, relevant preliminaries on CNN-based models are briefly introduced. In Section 3, we show the details of proposed deep learning models, including the construction of proposed network structures and the implementation of the recognition procedures. Section 4 demonstrates proposed methods on the TEP dataset and presents the application results. Finally, the conclusions are given in Section 5.

\section{Convolutional neural network}
\label{section2}

Aiming at realizing the known health state recognition of systems, CNN-based DL methods have been widely developed. In the existing publications, residual learning and multi-scale learning have become the main strategies to improve the feature extraction ability of CNNs.

\subsection{Standard learning}

CNN has shown outstanding performance in feature extraction and health states recognition tasks, relying on the unique characteristics of local receptive fields, spatial sub-sampling and weight sharing. Generally, the basic structure of a typical CNN includes an input layer, several alternately stacked convolutional and pooling layers, fully connected layers, and an output layer. The task of CNN as a feature extractor is accomplished by the forward propagation process.

In forward propagation, the entire network model extracts spatial information from the preprocessed input. As an important part of this process, convolutional layers utilize a series of convolutional kernels as filters to generate feature maps after sliding on the whole input neurons at the previous layer \citep{20182305285038}. The pooling layer performs scaled mapping immediately after each convolutional layer to reduce data dimensions and network parameters. Max-pooling, average-pooling, and stochastic-pooling are common methods for generating feature maps with low resolution \citep{JING20171}. After some stacked convolutional layers and pooling layers, the fully connected layer integrates the discriminative features from the previous layer into different classes.

\subsection{Residual learning}

In general, CNNs with deeper network structures are capable of extracting complex features, which can improve the performance of health states recognition. However, related studies have shown that when the number of network layers increases, the phenomenon of accuracy saturation and model performance degradation occurs, and leads to higher training errors \citep{he2015convolutional}. Residual learning is proposed as a solution to the problem that deeper networks cannot be easily trained \citep{he2016deep}. The basic concept of residual learning is that it is easier for a network to learn its residuals than it is to use several network layers to fit a hidden non-linear map \citep{yao2022adaptive}. 

Residual learning is achieved by constructing residual blocks, which typically consist of several convolutional layers, batch normalizations (BNs), ReLU activation functions, and one identity shortcut \citep{zhao2017deep}. CNNs incorporating residual learning are characterized by easy optimization, and the classification accuracy can be improved by increasing the depth of the network structure.

\subsection{Multi-scale learning}

Due to the fixed kernel scale in each convolutional layer, the multi-scale feature extraction capability of CNNs is limited. It is known that the small-size convolution kernel focuses on the details of the signal, while the large-size convolution kernel is responsible for capturing the characteristics that reflect the overall trend of the distribution. The information contained in the complementary features from different scales can help to better express health state information and improve the discrimination ability. The multi-scale learning strategy was developed to capture local features at different scales from the input measurement signals \citep{liu2019multiscale}.  

The multi-scale convolutional architectures can be realized by feature extraction in parallel through multi-branch CNNs with different convolution kernel sizes. CNNs with multi-scale learning can provide feature representations that contain complementary and rich state information compared with the single-scale model.

\subsection{CNN's general framework}

The Softmax layer is usually designed in the output layer as a classifier after CNN-based feature extractors to calculate the probability distribution of samples belonging to different classes. Consider a training data matrix $z_{tr}^{(i)}\in R^{w\times m}$, and corresponding training label $y_{tr}^{(i)}$, where $w$ and $m$ respectively represent the number of samples and variables in a training data matrix. The value of $y_{tr}^{(i)}$ is between 1 and $K$, where $K$ is the number of known health states. The Softmax layer outputs the conditional probability for class $k$:
\begin{align}
P(\hat{y}^{(k)}|l^{(1)}, \dots, l^{(K)})=\frac{{\rm exp}(l^{(k)})}{\sum_{k=1}^K {\rm exp}(l^{(k)})},
\end{align}
where the sum of the probabilities of all classes for sample $z_{tr}^{(i)}$ is 1. A CNN-based network with Softmax classifier is trained to increase the value $l^{(k)}$ of the target class $k$ relative to the other classes. The samples will be considered to belong to the predicted class with relatively large probability.

\begin{figure*}[!t]
	\centering
	\includegraphics[width=\hsize]{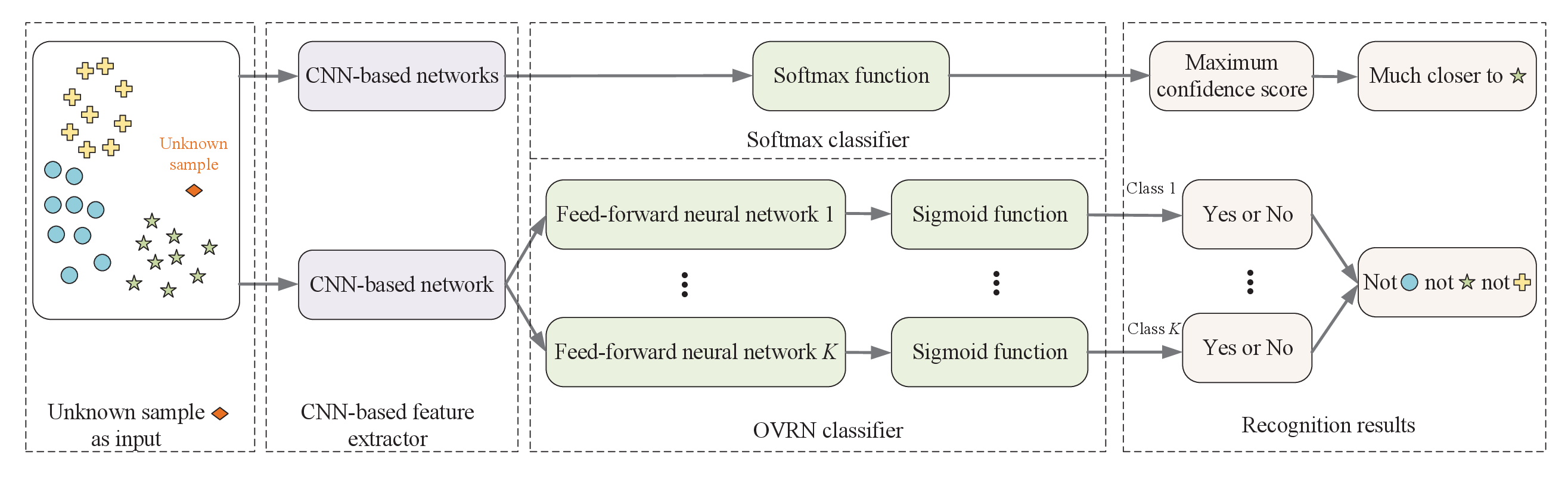}
	\caption{The general framework with Softmax classifier and new framework with OVRN classifier for recognition.}
	\label{ovrn}
\end{figure*}

For unknown health states recognition tasks, the mechanism of the general CNN-based framework with Softmax is shown in the upper part of Figure \ref{ovrn}. Since the Softmax function is designed to measure the relative likelihood of a known class compared to other known classes, unknown samples are easily assigned to the most similar known class with the maximum confidence score. Therefore, the unknown sample in Figure \ref{ovrn} is likely to be classified into the pentagram class that is closer than other classes, which is unfavorable for unknown states recognition.

\section{The proposed DL-based framework to deal with unknown states}
\label{section3}

Based on the advantages of OVRN with the learning ability of state specific discrimination features, a new DL-based framework incorporating OVRN for CNNs is proposed to be applied to industrial process health states recognition.

\subsection{The new OVRN based framework}

The recognition framework that introduces the One-vs-Rest Network (see the lower part of Figure \ref{ovrn}) gives a new solution to assign samples of unknown health states. For each known states, OVRN classifier is independently trained to learn specific features and provide a binary response, which indicates whether the sample belongs to this known state. Due to multiple low posterior probabilities of Sigmoid layers outputs, the unknown sample may be indicated not to belong to any known state.

\begin{figure*}[!t]
	\centering
	\includegraphics[width=\hsize]{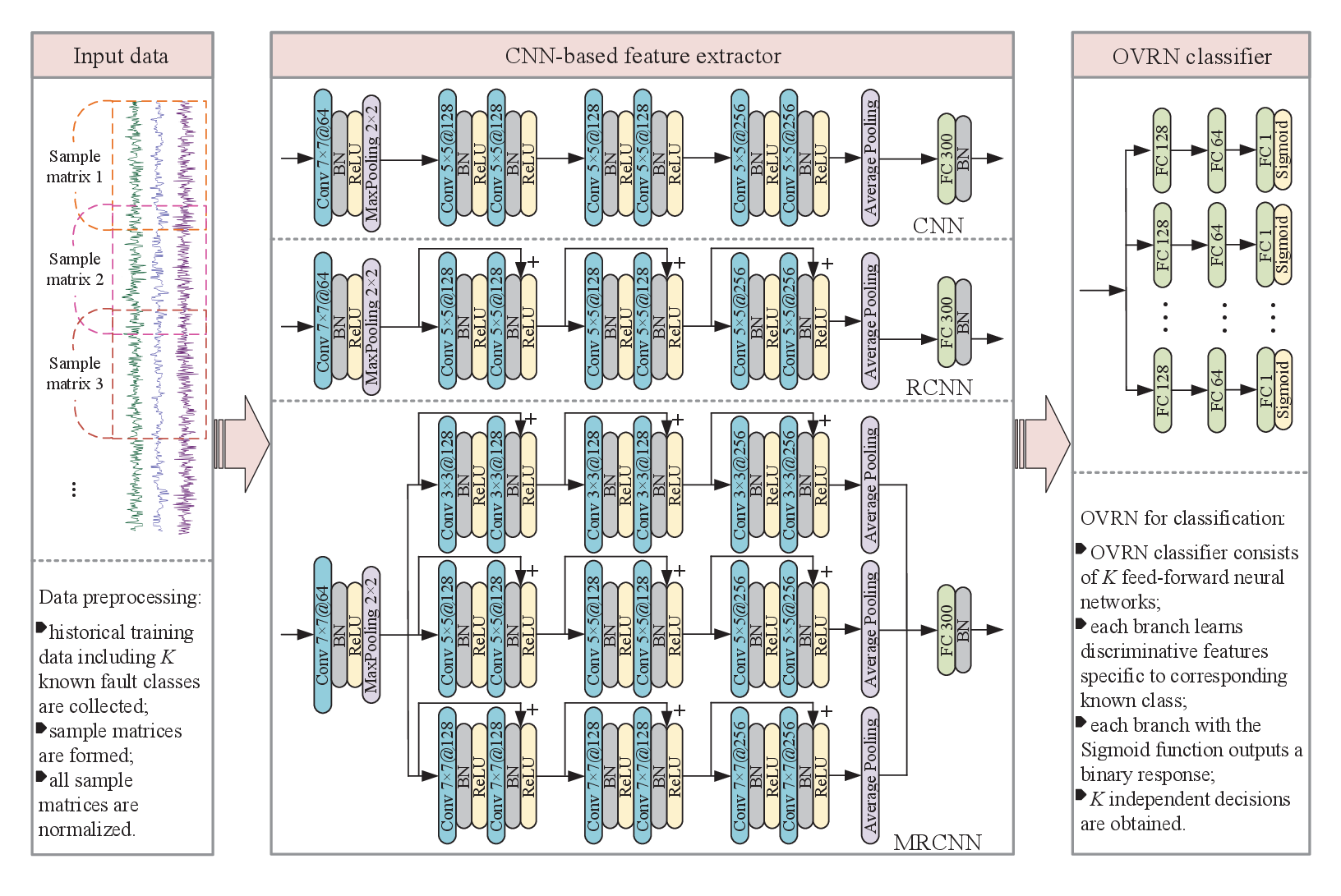}
	\caption{Architecture of the proposed framework.}
	\label{frame}
\end{figure*}

The proposed OVRN based framework consists of a CNN-based feature extractor and an OVRN classifier, as shown in Figure \ref{frame}. In this work, conventional CNN, residual CNN (RCNN) and multi-scale residual CNN (MRCNN) are employed as feature extractors.

Let $\mathcal{F}$ be the feature extractor, which can be one of traditional CNN, RCNN, and MRCNN. And $\mathcal{G}_1, \mathcal{G}_2, \dots, \mathcal{G}_K$ be $K$ parallel branch networks of OVRN. OVRN can better learn the discrimination information for target classes, and the output probability of $k$-th Sigmoid layer $\hat{y}^{(k)}$ for class $k$ can be expressed as:
\begin{align}
P(\hat{y}^{(k)}|z_{tr}^{(i)})=\mathcal{G}_k(\mathcal{F}(z_{tr}^{(i)})).
\label{eq1}
\end{align}
Since the Sigmoid function is applied to node $l^{(k)}$ in last output layer, the relationship between the probability $P(\hat{y}^{(k)}|z_{tr}^{(i)})$ in \eqref{eq1} and the output of node $l^{(k)}$ can be expressed as:
\begin{align}
P(\hat{y}^{(k)}|l^{(k)})=\frac{1}{1+{\rm exp}(-l^{(k)})}.
\end{align}
Obviously, each output layer with Sigmoid activation function provides independent decision to distinguish whether samples belong to this class or not. Therefore, OVRN using Sigmoid layers can provide more supporting information for unknown health states recognition, compared with traditional Softmax layer.

In the recognition framework using OVRN as the classifier, the loss function for model training is expressed as the sum of $K$ binary cross-entropy losses as follows \citep{jang2021collective}:
\begin{align}
\mathcal{L}_{tr}=&-\frac{1}{N}\sum^{N}_{i=1} \sum^{K}_{k=1}(\mathbb{I}(y_{tr}^{(i)}=\hat{y}^{(k)})P(\hat{y}^{(k)}|z_{tr}^{(i)})\\ \nonumber
&+\mathbb{I}(y_{tr}^{(i)} \neq \hat{y}^{(k)})(1-P(\hat{y}^{(k)}|z_{tr}^{(i)})),
\end{align}
where $N$ is the batch size and $\mathbb{I}$ represents the indicator function. During training, this loss function allows each branch network of OVRN to learn relevant feature representations that contribute to its binary recognition task. The parameter set $\theta=\{\theta_{\mathcal{F}}, \theta_{\mathcal{G}_1}, \theta_{\mathcal{G}_2}, \dots ,\theta_{\mathcal{G}_K}\}$ can be updated through the gradient descent algorithm to minimize the loss as the optimization object. $\theta_{\mathcal{F}}, \theta_{\mathcal{G}_1}, \theta_{\mathcal{G}_2}, \dots ,\theta_{\mathcal{G}_K}$ are the required parameters of the feature extractor and the OVRN branch networks.

\subsection{Recognition rule based on collective decisions}

The collective decision score for state $k$ is computed, given the sample matrix $z_{tr}^{(i)}$, based on the logit of OVRN outputs $P(\hat{y}^{(1)}|z_{tr}^{(i)}), P(\hat{y}^{(2)}|z_{tr}^{(i)}), \dots, P(\hat{y}^{(K)}|z_{tr}^{(i)})$, and is expressed as \citep{jang2021collective}:
\begin{align}
S_{cd}^{i,k}=&{\rm log} (P(\hat{y}^{(k)}|z_{tr}^{(i)}))\\ \nonumber
&-\frac{1}{K-1}\sum_{j \neq k} {\rm log}(P(\hat{y}^{(j)}|z_{tr}^{(i)})).
\end{align}

Therefore, samples belonging to one of the known states receive higher scores for the target state and lower scores for other states. 

Since $K$ collective decision scores are obtained for the sample matrix $z_{tr}^{(i)}$, a collective decision rule is proposed. Considering samples from unknown health states, the rule is expressed as:
\begin{gather}
z_{tr}^{(i)} \in \text{State} \; k^*,\; \text{if} \; S_{cd}^{i,k^*} \ge \varepsilon_{k^*} ,\\ \nonumber
\text{subject to} \; k^* = \mathop{\arg\max}_{k=1,\dots,K} S_{cd}^{i,k},
\end{gather}
where $\varepsilon_k$ is the collective decision score threshold for health states $k$, and it can be obtained by ensuring that 95\% of the training samples from class $k$ can be diagnosed as the target class based on their collective decision scores. Under this rule, samples are recognized as belonging to the known state with the largest collective decision score value above the corresponding threshold. Otherwise, the samples are assigned to the unknown health states.

\subsection{Outline of the proposed recognition strategy}

Conventional CNN-based models applied to predictive maintenance can efficiently perform recognition tasks under the assumption that labeled samples are sufficient. However, these methods fail to assign samples from unknown health states that did not participate in the training process. This work allows CNN-based models to overcome their limitations in applications where the classes of test samples are unknown. The proposed recognition framework has been applied to three CNN-based models including conventional CNN, RCNN, and MRCNN. We define the these three improved CNN-based models for unknown health states recognition as CNN-OVRN, RNN-OVRN and MRCNN-OVRN. Compared respectively to CNN, RCNN and MRCNN, the proposed methods can simultaneously achieve classification of known and unknown health states. This good performance benefits from the enhanced ability of OVRN to reject unknown samples and the complex decision boundaries established by the collective decision rule. In addition, the classification performance of the proposed framework to recognize known health states can also be guaranteed, as shown for instance below by CNN-OVRN, RNN-OVRN and MRCNN-OVRN.
\begin{algorithm}[!t]
	\renewcommand{\algorithmicrequire}{\textbf{Input:}}
	\renewcommand{\algorithmicensure}{\textbf{Output:}}
	\caption{Unknown health states recognition with
		collective decision based DL networks } 
	\label{alg}
	\begin{algorithmic}
		\begin{spacing}{1.2}
			\Require 
			training data set $X_{tr}=\{ x_{tr}^{(1)}, \dots, x_{tr}^{n_{tr}}\}$, $X_{tr} \in R^{n_{tr}\times m}$, and training label set, $Y_{tr}\in R^{n_{tr}}$; testing data set $X_{te}=\{ x_{te}^{(1)}, \dots, x_{te}^{n_{te}}\} $, $X_{te} \in R^{n_{te}\times m}$; 
			\Ensure
			predicted healthy state
			\State $Z_{tr}\in R^{(n_{tr}-w+1)\times m}, Z_{te}\in R^{(n_{te}-w+1)\times m} $ $\gets$ data pre-processing  
			\State $\triangleright$ Training a CNN-based model
			\State $\theta=\{\theta_{\mathcal{F}}, \theta_{\mathcal{G}_1}, \theta_{\mathcal{G}_2}, \dots ,\theta_{\mathcal{G}_K}\}$ $\gets$ initialize parameters
			\Repeat
			\State $\mathcal{L}_{tr}=-\frac{1}{N}\sum^{N}_{i=1} \sum^{K}_{k=1}(\mathbb{I}(y_{tr}^{(i)}=\hat{y}^{(k)})P(\hat{y}^{(k)}|z_{tr}^{(i)})+\mathbb{I}(y_{tr}^{(i)} \neq \hat{y}^{(k)})(1-P(\hat{y}^{(k)}|z_{tr}^{(i)}))$
			\State $\theta=\{\theta_{\mathcal{F}}, \theta_{\mathcal{G}_1}, \theta_{\mathcal{G}_2}, \dots ,\theta_{\mathcal{G}_K}\}$ $\gets$ updata parameters using gradients of $\mathcal{L}$
			\Until{the convergence of loss $\mathcal{L}_{tr}$}
			\State  $\triangleright$ Training the collective decision rule
			\For {$\boldsymbol{i=1}$ \textbf{to} $\boldsymbol{n_{tr}}$}
			\State $S_{cd}^{i,k}={\rm log} (P(\hat{y}^{(k)}|z_{tr}^{(i)}))-\frac{1}{K-1}\sum_{j \neq k} {\rm log}(P(\hat{y}^{(j)}|z_{tr}^{(i)}))$ $\gets$ decision score
			\EndFor
			\State $\varepsilon_1, \varepsilon_2, \dots, \varepsilon_K$ $\gets$ collective decision score thresholds for Fault 1, 2, $\dots$, $K$
			\State $\triangleright$ Online recognition procedure
			\For {$\boldsymbol{i=1}$ \textbf{to} $\boldsymbol{n_{te}}$}
			\State  $S_{cd}^{i,k}, k=1,\dots,k$ $\gets$ collective decision scores of all health states
			\State $k^* = \underset{k=1, \cdots, K}{\arg \max } S_{cd}^{i,k} $ $\gets$ Fault $k^*$ with the largest collective decision scores
			\If{$S_{cd}^{i,k^*}\geq \varepsilon_{k^*} $}
			\State $z_{te}^{(i)}$ is classified as state ${k^*}$
			\Else
			\State $z_{te}^{(i)}$ is recognized as unknown health states
			\EndIf
			\EndFor
		\end{spacing}
	\end{algorithmic}
\end{algorithm}

The training and recognition process of the proposed framework are shown in Algorithm \ref{alg}. The whole procedure can be divided into two stages: offline stage and online stage, which can be summarized as follows.
\begin{itemize}
	\item Offline stage:
	\begin{itemize}
		\item Data from training set are pre-processed. The samples in the adjacent $w$ sampling windows are formed into a sample matrix with the size of $w\times m$, and then normalized to zero mean and one deviation.
		\item The model is trained to obtain the optimal parameters. The training process stops when the loss $\mathcal{L}_{tr}$ of the training set tends to converge.
		\item The collective decision rule is trained. The collective decision scores of training sample matrices are first calculated. The threshold $\varepsilon_k$ is then set by ensuring that 95\% of the Fault $k$ sample matrices are assigned to the target health state.
	\end{itemize}
	\item Online stage:
	\begin{itemize}
		\item The test data were preprocessed using the same sampling window and normalization parameters as the training set.
		\item  The output values of the model are calculated, and the decision scores corresponding to all health states are further obtained.
		\item The health state corresponding to the test data are predicted according to the collective decision rule.
	\end{itemize}
\end{itemize}

\section{Application on Tennessee Eastman process}
\label{section4}

\subsection{Experimental Settings}

Tennessee Eastman process (TEP) is a simulation program developed from a actual chemical plant, which has been widely employed in the verification of various health states recognition methods \cite{YU20191}. The additional TEP simulation datasets for anomaly evaluation was published online by Rieth et al. \cite{DVN/6C3JR1_2017}, which are adopted for training and testing of the proposed method. Both the training and test sets provide data of 500 random simulation runs for 20 health state types. For the training and test sets, each simulation run lasts 25 hours and 48 hours with a sampling time of 3 minutes, generating 500 and 960 samples. A total of 52 variables, including 41 measured variables and 11 manipulated variables, are employed for health states recognition. 

From the training dataset, 15 simulation runs are used for training. 1 simulation run from the test set is used as test data, which included 15 known and 5 unknown health states. After the samples for training and testing are determined, the adjacent 20 samples (1 hour) are formed into a sample matrix with a size of $20\times 52$. To realize the validation of the proposed method in known and unknown health states recognition tasks, the required data are set as shown in Table \ref{tedata}. 

\begin{table*}[!ht]
	\renewcommand\arraystretch{1.2}
	\centering
	\caption{TEP datasets}
	\label{tedata}
	\resizebox{\textwidth}{!}{
	\begin{tabular}{llll}
		\hline
		& Health state & Number of samples & Number of sample matrices   \\ \hline
		Training & IDV 1-18 (except for IDV 3, 9, 15) & 500$\times$ 15 $\times$ 15 & 481 $\times$ 15 $\times$ 15 \\ 
		Testing (known) & IDV 1-18 (except for IDV 3, 9, 15) & 960$\times$ 15 $\times$ 1 & 941 $\times$ 15 $\times$ 1 \\ 
		Testing (Unknown) & IDV 3, 9, 15, 19, 20 & 960$\times$ 5 $\times$ 1 & 941 $\times$ 5 $\times$ 1 \\ 
		\hline
	\end{tabular}
    }
\end{table*}

A new deep learning framework based on CNN-based feature extractors and OVRN classifier is proposed for known and unknown health states recognition. This work allows CNN-based models such as conventional CNN, RCNN, and MRCNN to overcome limitations when faced with new observations that are not part of the training set. The network structures adopted for CNN-OVRN, RCNN-OVRN and MRCNN-OVRN are shown in Figure \ref{frame}. All deep learning models in our experiments were trained on a NVIDIA GeForce RTX 2080 Ti GPU and the Adam optimizer with a learning rate of 0.0005.

\subsection{Methods Comparison}

When the proposed CNN-OVRN, RCNN-OVRN, and MRCNN-OVRN perform online recognition, the samples are labeled as one of the known health states or the unknown health state (UHS). Therefore, for a comprehensive comparison with other related models, this work evaluates the recognition performance of the proposed framework on both known and unknown states. The quantitative metrics used for evaluation mainly include: recognition accuracy of known states, recognition accuracy of unknown states, and overall recognition accuracy.

To verify the recognition effectiveness of the proposed methods, statistical analysis-based models and CNN-based baseline networks from related literature are compared according to the above evaluation metrics. 
\begin{itemize}
	\item Statistical analysis-based models: discriminant analysis methods show inherent superiority in health states recognition tasks based on the assumption that the training dataset is available and complete. The existing literature has proposed improved discriminant rules for common discriminant analysis methods such as quadratic discriminant analysis (QDA), Fisher discriminant analysis (FDA), and kernel Fisher discriminant (KFD) \cite{lou2022novel}. The improved discriminant analysis methods are able to handle observations belonging to new health states, named UQDA, UFDA and UKFD.
	\item CNN-based baseline networks: the proposed CNN-OVRN, RCNN-OVRN and MRCNN-OVRN are first completed for fault recognition on the test dataset. MCNN-OVRN is constructed to verify the mitigation of residual learning for performance degradation, where the multi-scale RCNN is replaced by a multi-scale CNN with the same network structure. To illustrate the contribution of the collective decision rule in proposed methods, deep learning models of MRCNN-Softmax and MRCNN-OVRN without collective decision rules are used for comparison. MRCNN-Softmax employs the Softmax layer instead of OVRN classifier for decision making. Both models assign samples to known states with the largest output value greater than a pre-set threshold, otherwise, are considered to be unknown faults. Here, the threshold is set to 0.5, which is a common value in unknown state recognition \cite{jang2021collective}.
\end{itemize}

\subsection{Health states recognition result and performance comparison}

Considering the test data in Table \ref{tedata}, the health states recognition accuracies of proposed methods and related comparative models for fifteen known and unknown states are shown in Table \ref{result}. For each experimental result, the accuracies from ten independent runs of the deep learning-based models were averaged to avoid random interference. It can be observed that UFDA, UQDA and UKFD cannot achieve good recognition results for some health states such as states 6, 16, 17 and 18. Also, the recognition accuracy of these three methods for samples of unknown states is quite low compared to deep learning based models. The reasons why discriminant analysis methods perform poorly on both known and unknown recognition tasks can be explained from two perspectives. One is that the limited learning ability of these shallow learning-based methods makes it impossible to guarantee that the extracted features can best represent the discriminative information of health states. Second, the complex mapping relationship between input data and health states may be difficult to be fully learned due to the independent execution of feature extraction and health states recognition.

\begin{table*}[!ht]
	\centering
	\caption{The comparison of fault classification results.}
	\label{result}
	\begin{threeparttable}
		\resizebox{\textwidth}{!}{
		\begin{tabular}{p{37pt}p{30pt}p{30pt}p{30pt}p{30pt}p{30pt}p{30pt}p{30pt}p{30pt}p{30pt}p{30pt}p{30pt}}
			\hline
			Item & (a)  & (b)   & (c) & (d)  &  (e)  & (f) & (g) & (h) & (i) & (j) & (k) \\
			\hline
			IDV 1  & 0.8    & 0.8094 & 0.8    & 0.8087 & 0.8232 & 0.8200 & 0.8232 & 0.8165 & 0.8393 & 0.8413 & 0.8253 \\
			IDV 2  & 0.8125 & 0.8167 & 0.8115 & 0.8082 & 0.8272 & 0.8196 & 0.8183  & 0.8231 & 0.8463 & 0.8400 & 0.8244 \\
			IDV 4  & 0.8    & 0.8135 & 0.8063 & 0.7991 & 0.7985 & 0.8032 & 0.8094 & 0.7835 & 0.8346 & 0.8321 & 0.8083 \\
			IDV 5  & 0.7167 & 0.8135 & 0.7740 & 0.8009 & 0.7902 & 0.8043 & 0.7815 & 0.7932 & 0.8380 & 0.8394 & 0.8043 \\
			IDV 6  & 0.6208 & 0.4781 & 0.3854 & 0.8234 & 0.8298 & 0.8395 & 0.8402 & 0.8257& 0.8474 & 0.8478 & 0.8356 \\
			IDV 7  & 0.8208 & 0.8146 & 0.7865 & 0.7829 & 0.7756 & 0.7926 & 0.7836 & 0.7865 & 0.8476 & 0.8493 & 0.7909 \\
			IDV 8  & 0.7083 & 0.7208 & 0.5896 & 0.7122 & 0.7256 & 0.7046 & 0.7046 & 0.7301 & 0.7868 & 0.8046 & 0.7387 \\
			IDV 10 & 0.65   & 0.8354 & 0.7865 & 0.7538 & 0.7547 & 0.7411 & 0.7352 & 0.7595 & 0.7879 & 0.7931 & 0.7498 \\
			IDV 11 & 0.4740 & 0.8208 & 0.675 & 0.7825 & 0.7953 & 0.7872 & 0.7870 & 0.7826 & 0.8295 & 0.8336 & 0.7988 \\
			IDV 12 & 0.6792 & 0.7906 & 0.6167 & 0.8045 & 0.8136 & 0.8081 & 0.8102 & 0.8099 & 0.8399 & 0.8360 & 0.8104 \\
			IDV 13 & 0.6552 & 0.7583 & 0.4479 & 0.7758 & 0.8049 & 0.8102 & 0.8091 & 0.7805 & 0.7613 & 0.8064 & 0.8130 \\
			IDV 14 & 0.6865 & 0.8198 & 0.8052 & 0.8062 & 0.8149 & 0.8172 & 0.8074 & 0.8081 & 0.8457 & 0.8459 & 0.8157 \\
			IDV 16 & 0.6552 & 0.7521 & 0.7313 & 0.8266 & 0.8544 & 0.8461 & 0.8308 & 0.8471 & 0.9245 & 0.9078 & 0.8168 \\
			IDV 17 & 0.5042 & 0.6448 & 0.5010 & 0.7763 & 0.7751 & 0.7743 & 0.7749 & 0.7761 & 0.7751 & 0.7747 & 0.7730 \\
			IDV 18 & 0.2698 & 0.2052 & 0.1552 & 0.7937 & 0.7962 & 0.7953 & 0.7974 & 0.7963 & 0.7951 & 0.7974 & 0.7960 \\
			UFC    & 0.0902 & 0.1643 & 0.1075 & 0.6163 & 0.6556 & 0.6997 & 0.6786 & 0.6241 & 0.2965 & 0.3180 & 0.7194 \\
			\hline
		\end{tabular}
	}
		\scriptsize		
		(a): UFDA; (b): UQDA; (c): UKFD; (d): CNN-OVRN with kernel size of $5\times 5$; (e): RCNN-OVRN with kernel size of $3\times 3$; (f): RCNN-OVRN with kernel size of $5\times 5$; (g): RCNN-OVRN with kernel size of $7\times 7$; (h): MCNN-OVRN; (i): MRCNN-Softmax; (j): MRCNN-OVRN without the collective decision rule; (k): MRCNN-OVRN.
	\end{threeparttable}
\end{table*} 

The results in Table \ref{result} show that the proposed methods have superior recognition performance for both known and new health states. On the one hand, the proposed methods achieve satisfactory single-recognition accuracy relying on strong nonlinear feature extraction capabilities and an end-to-end recognition framework. On the other hand, observations belonging to new health states have been well identified by the proposed methods, as demonstrated by the recognition accuracy of CNN-OVRN, RCNN-OVRN and MRCNN-OVRN for unknown states all maintained above 0.6. Due to the advantages of residual learning and multi-scale learning, MRCNN-OVRN perform best with a recognition accuracy of 0.7194. As a comparison method, the recognition accuracy of MCNN-OVRN for unknown states is 0.6241 lower than that of MRCNN-OVRN with the same network structure except that there is no residual learning. It can be explained that the degradation problem caused by the deep network structure may limit the feature learning ability so that the distinguishability between known and unknown states is slightly poor. The results of MRCNN-Softmax and MRCNN-OVRN without the collective decision rule are to demonstrate that the decision rule of the proposed method is capable of providing robust recognition results. These two methods slightly improve the recognition accuracy for known states due to loose decision boundaries constructed, however, the recognition accuracy for unknown states is only around 0.3.

\begin{figure*}[!t]
	\centering
	\includegraphics[width=\hsize]{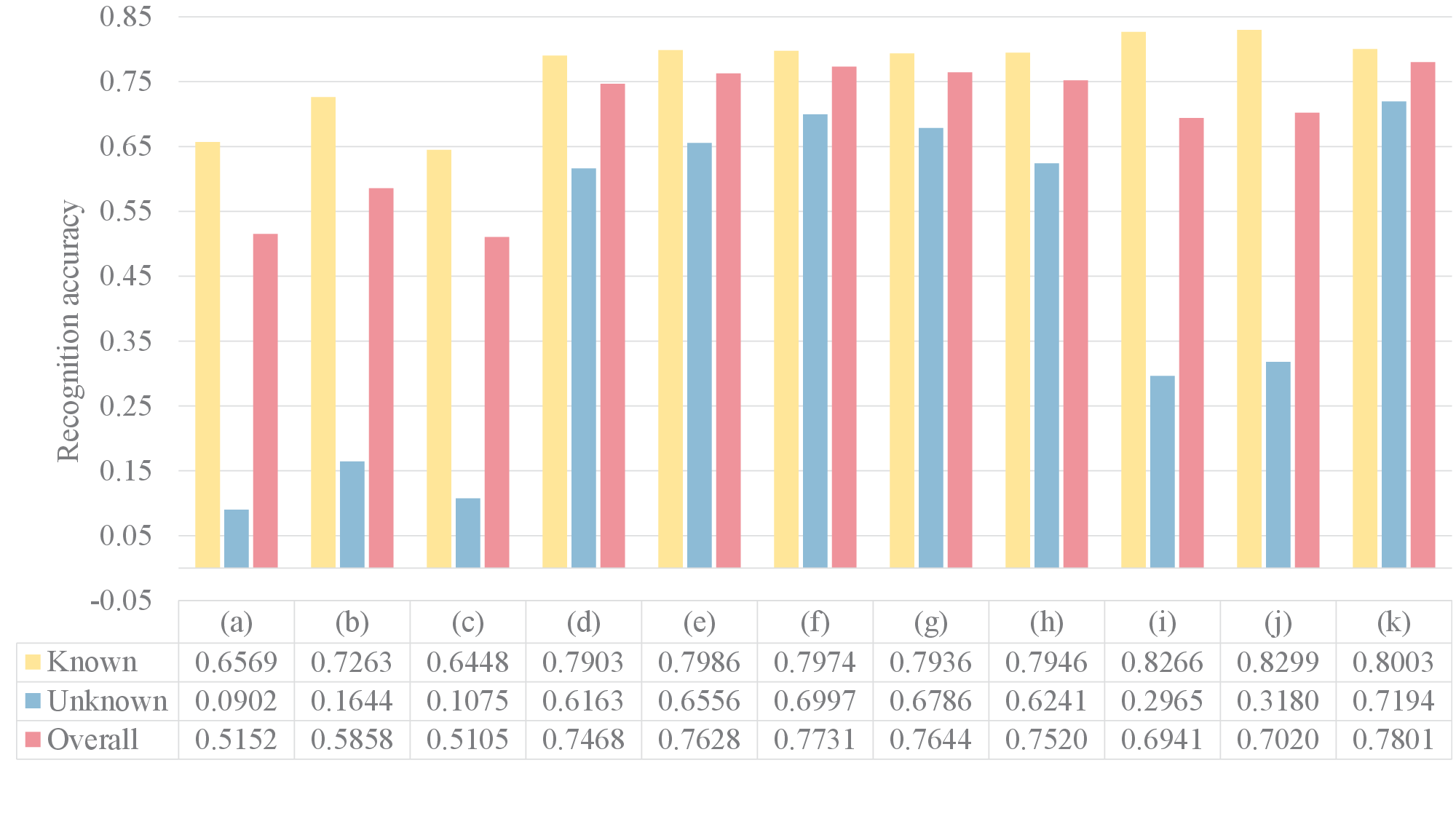}
	\caption{The comparison of health states recognition results.}
	\label{TEresult}
\end{figure*}

Fig. \ref{TEresult} provides the recognition accuracy of all methods on the same test set, including the average recognition accuracy of known states, the recognition accuracy of unknown states, and the overall accuracy.  The recognition results of unknown states prove that the online recognition performance of the proposed methods is superior. Although there is a small reduction in the accuracy of known states compared to models that do not employ collective decision rules, the recognition performance of the proposed method is within an acceptable range. The overall recognition accuracy of CNN-OVRN, RCNN-OVRN and MRCNN-OVRN are all higher than 0.74, and MRCNN-OVRN benefits from multi-scale residual learning to improve the accuracy to 0.7801.

\begin{figure*}[!t]
	\centering
	\includegraphics[width=\hsize]{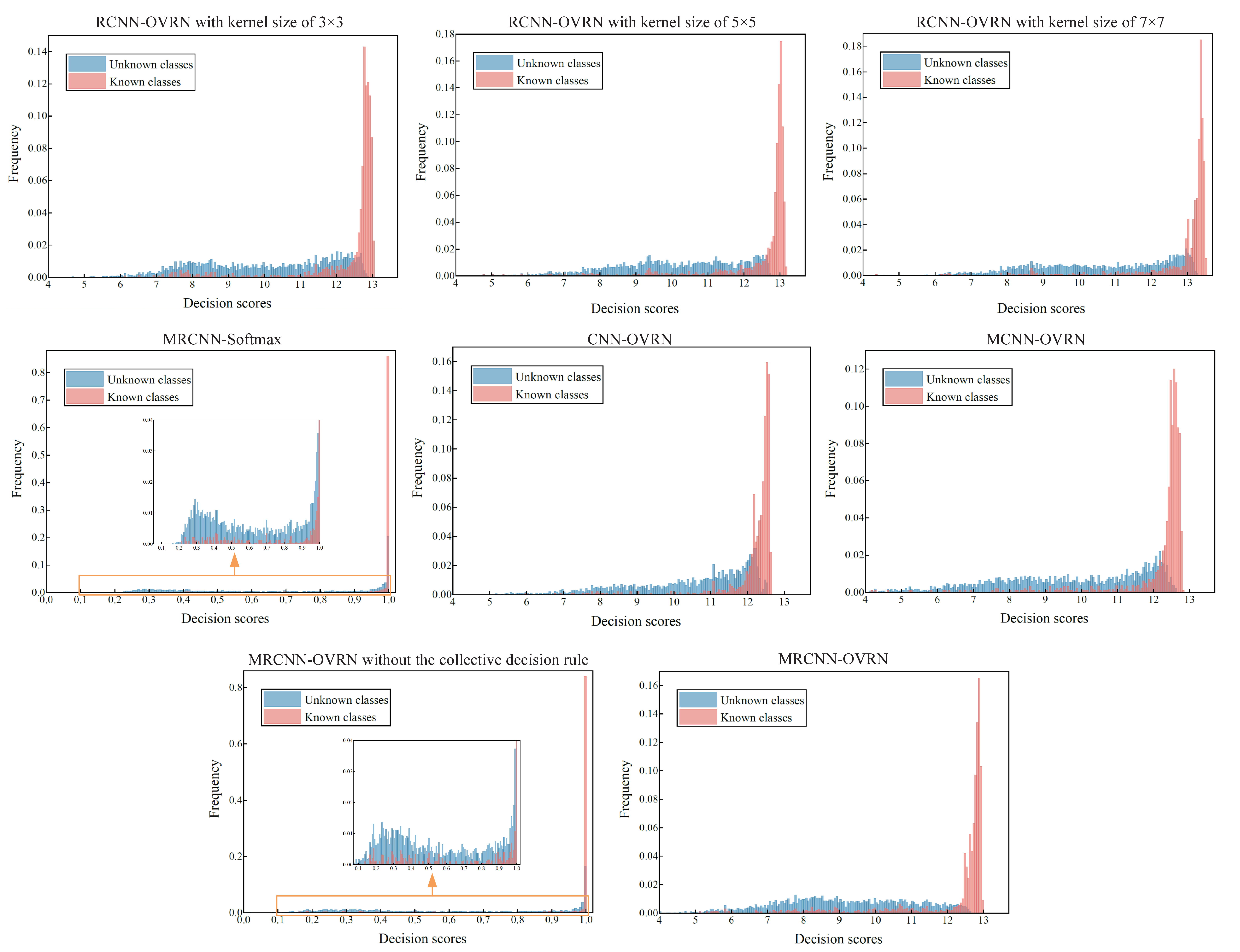}
	\caption{Distribution of decision scores for known and unknown health states on TEP.}
	\label{distribution}
\end{figure*}

Further, we demonstrate the effectiveness of the proposed methods by comparing the distribution histograms of decision scores for known and unknown health states, as shown in Figure \ref{distribution}. Greater separation between the histograms of known states (red part) and unknown states (blue part) indicates a higher recognition ability for unknown states. CNN-OVRN, RCNN-OVRN and MRCNN-OVRN can obtain clear separation results between samples of known and unknown states with small overlap due to the collaboration of CNN-based feature extractors and OVRN classifer. MRCNN-OVRN provides the best separation histogram, which is reflected in the fact that the decision scores of samples from unknown state are concentrated around small values. Compared with MRCNN-OVRN, MCNN-OVRN without residual learning has a larger overlapping area of the distribution of decision scores, which limits the learning ability of deeper features. For MRCNN-Softmax and MRCNN-OVRN without the collective decision rule, the maximum posterior probability output by Softmax layer and Sigmoid layer is taken as the decision score, respectively. The Softmax layer, which provides high confidence scores for unknown samples, results in poor distinguishability of known and unknown states by identifying the state with the highest relative likelihood among all known states. MRCNN-OVRN without the collective decision rule provides slightly lower decision scores than MRCNN-Softmax since each node of the output layer provides an independent decision to distinguish whether a sample belongs to this state or not.

\begin{figure*}[!ht]
	\centering
	\includegraphics[width=\hsize]{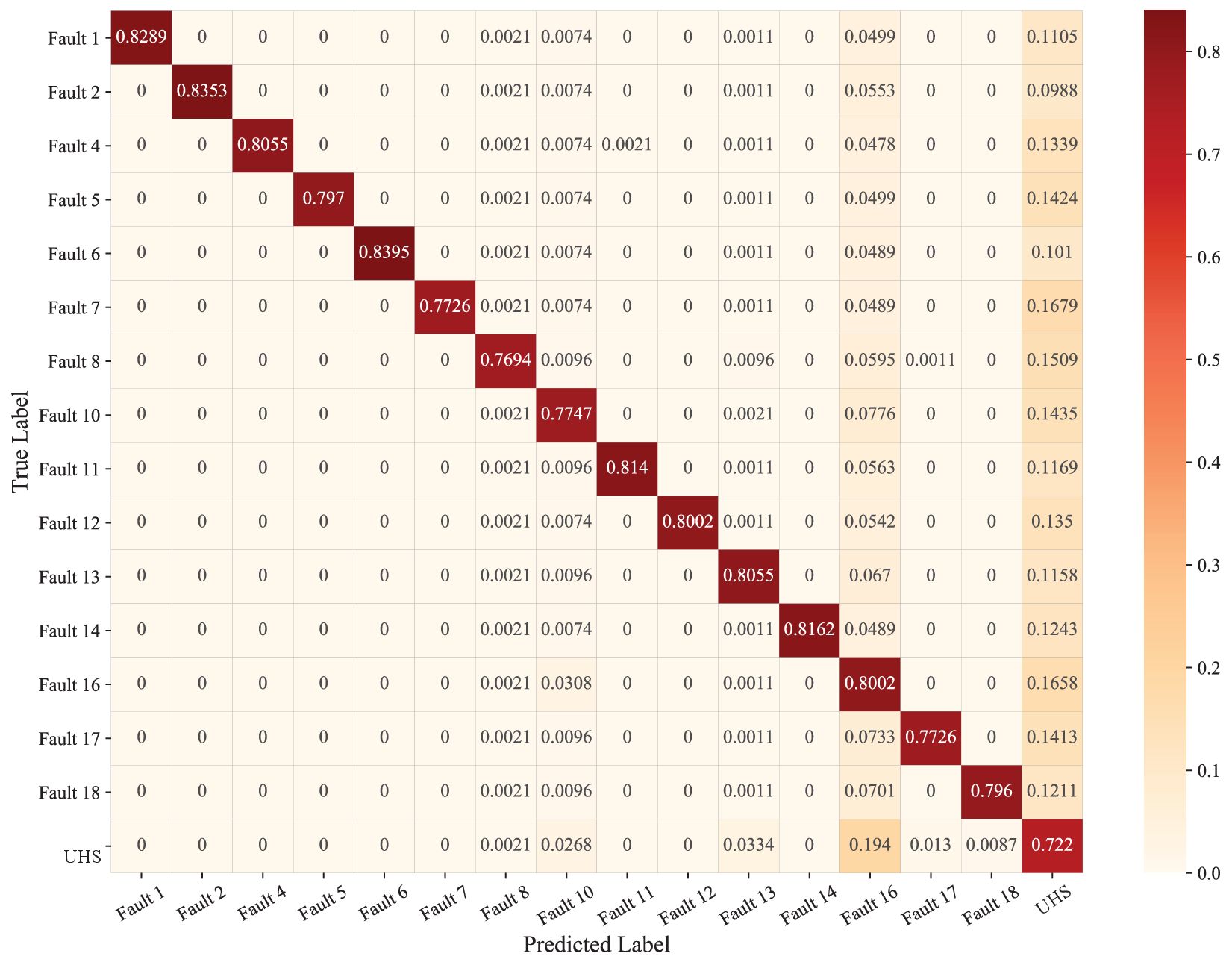}
	\caption{Confusion matrix of MRCNN-OVRN.}
	\label{matrix}
\end{figure*}

For one of ten experimental runs, Figure \ref{matrix} shows the confusion matrix for test samples of 15 known health states and the unknown state, where the rows and columns represent true and predicted states. In this confusion matrix, one can clearly see how all the test data are assigned by MRCNN-OVRN. It can be observed that MRCNN-OVRN can achieve the recognition of most unknown samples while maintaining satisfactory recognition performance for known states. This is mainly due to the multi-scale feature learning ability of MRCNN and the complex decision boundary established by the OVRN classifier. It can also be found that a small number of samples of known states are misassigned to states 8, 10, 13 and 16. In addition, there are also samples that belong to known health states that are considered unknown. To enhance the discriminative performance for unknown states, the proposed method establishes a strict and compact decision boundary to reject new samples as unknown states. Therefore, under the collective decision rule, some samples of known states are inevitably misidentified as unknown states.

\section{Conclusion}
\label{section5}

In this article, a novel collective decision framework combining a CNN-based feature extractor and an OVRN classifier is proposed to address the important challenge of deep learning models facing new test samples. Our work extends the CNN-based health recognition schemes, for instance conventional CNN, residual CNN (RCNN), and multi-scale residual CNN (MRCNN). It allows these methods to simultaneously achieve recognition of known and unknown health states. CNN-based models learn abstract and generalized latent representations, while the features extracted by OVRN and decision boundaries established by the collective decision rule are beneficial for rejecting unknown health states and preserving satisfactory recognition performance for known states. The public dataset from Tennessee Eastman process (TEP) was carried out to validate the proposed methods. Specifically, the effectiveness of the proposed CNN-OVRN, RCNN-OVRN, and MRCNN-OVRN was validated by a comparison with several statistical analysis-based models and CNN-based baseline networks. The proposed methods are demonstrated to have satisfactory accuracy in both known and unknown states recognition by comparing the recognition accuracy and analyzing the distribution histogram of decision scores obtained by the test methods. The significance of this work was to provide a generalizable deep learning-based framework for states recognition of industrial processes under the assumption that not all health states are known. 

Further research of this work can be implemented from the following aspects. First, how to merge the identified samples from unknown health states into the original training dataset and how to retrain the existing model is a meaningful challenge. Second, the selection of hyperparameters relies on manual setting and repeated experiments, which can consume much time to obtain satisfactory recognition performance. It might be a possible development direction that allows the optimal usage of machine learning with deep learning models to realize feature learning in a fast and automatic way.

\bibliographystyle{model}
\bibliography{ref}

\end{document}